\documentclass{article}

\usepackage[final]{corl_2019} %

\usepackage[numbers]{natbib}
\usepackage{amsmath,amssymb,amsfonts}
\usepackage{algorithm}
\usepackage[noend]{algpseudocode}
\usepackage{graphicx}
\usepackage{color}
\usepackage[normalem]{ulem}
\usepackage{siunitx}
\usepackage{booktabs}
\usepackage{mathtools}
\usepackage{caption}
\usepackage{subcaption}
\DeclarePairedDelimiter\norm{\lVert}{\rVert}%
\DeclarePairedDelimiter\abs{\lvert}{\rvert}%

\DeclareMathOperator*{\argmin}{arg\,min}

\usepackage [english]{babel}
\usepackage [autostyle, english = american]{csquotes}
\MakeOuterQuote{"}

\makeatletter
\let\oldabs\abs
\def\abs{\@ifstar{\oldabs}{\oldabs*}}
\let\oldnorm\norm
\def\norm{\@ifstar{\oldnorm}{\oldnorm*}}
\makeatother

\usepackage[capitalise]{cleveref}

\newcommand{\papertitle}{TuneNet: One-Shot Residual Tuning for System Identification and Sim-to-Real Robot Task Transfer}
\title{\papertitle{}}

\author{
Adam Allevato\\
Department of Mechanical Engineering\\
The University of Texas at Austin\\
\texttt{allevato@utexas.edu}
\And
Elaine Schaertl Short\\
Department of Electrical and Computer Engineering\\
The University of Texas at Austin\\
\texttt{eshort@ece.utexas.edu}
\And
Mitch Pryor\\
Department of Mechanical Engineering\\
The University of Texas at Austin\\
\texttt{mpryor@utexas.edu}
\And
Andrea Thomaz\\
Department of Electrical and Computer Engineering\\
The University of Texas at Austin\\
\texttt{athomaz@ece.utexas.edu}
}

\begin{document}
\maketitle

\begin{abstract}
As researchers teach robots to perform more and more complex tasks, the need for realistic simulation environments is growing. Existing techniques for closing the reality gap by approximating real-world physics often require extensive real world data and/or thousands of simulation samples. This paper presents TuneNet, a new machine learning-based method to directly tune the parameters of one model to match another using an \textit{iterative residual tuning} technique. TuneNet estimates the parameter difference between two models using a single observation from the target and minimal simulation, allowing rapid, accurate and sample-efficient parameter estimation. The system can be trained via supervised learning over an auto-generated simulated dataset. We show that TuneNet can perform system identification, even when the true parameter values lie well outside the distribution seen during training, and demonstrate that simulators tuned with TuneNet outperform existing techniques for predicting rigid body motion. Finally, we show that our method can estimate real-world parameter values, allowing a robot to perform sim-to-real task transfer on a dynamic manipulation task unseen during training. Code and videos are available online at \url{http://bit.ly/2lf1bAw}.
\end{abstract}

\section{Introduction}
Recent research has validated simulators for real-world robot behaviors and control policies. Several studies have shown the ability to adapt simulation-learned policies to the real world based on observations \citep{Johannink_2018_residual, Chebotar2018SimToRealAdaptation, AAAI17-Hanna, golemo2018simrealtransfer, zeng_2019_tossingbot, silver_2018_residual, rusu2017sim}, but these techniques require extensive real-world data collection. Even before collecting data in the real world, however, we can improve simulations by simply selecting more realistic simulator parameter values. While some approaches train over a large set of possible parameter values to create a robust policy \citep{Tobin2017, tan_sim_to_real_2018, RajeswaranEPOpt, pinto_2018_asymmetric_rl, Chebotar2018SimToRealAdaptation}, we are interested in the problem of creating a single "canonical" simulation that behaves as realistically as possible, which can then be used for robust physics prediction or better task planning.

In this paper, we propose TuneNet, a \textit{residual tuning} technique that uses a neural network to modify the parameters of one physical model so it approximates another. TuneNet takes as input observations from two different models (i.e. a simulator and the real world), and estimates the difference in parameters between the models. By estimating the parameter gradient landscape, a small number of iterative tuning updates enable rapid convergence on improved parameters from a single observation from the target model. TuneNet is trained using supervised learning on a dataset of pairs of auto-generated simulated observations, which allows training to proceed without real-world data collection or labeling.

\begin{figure}[t]
    \centering
    \includegraphics[width=0.5\textwidth]{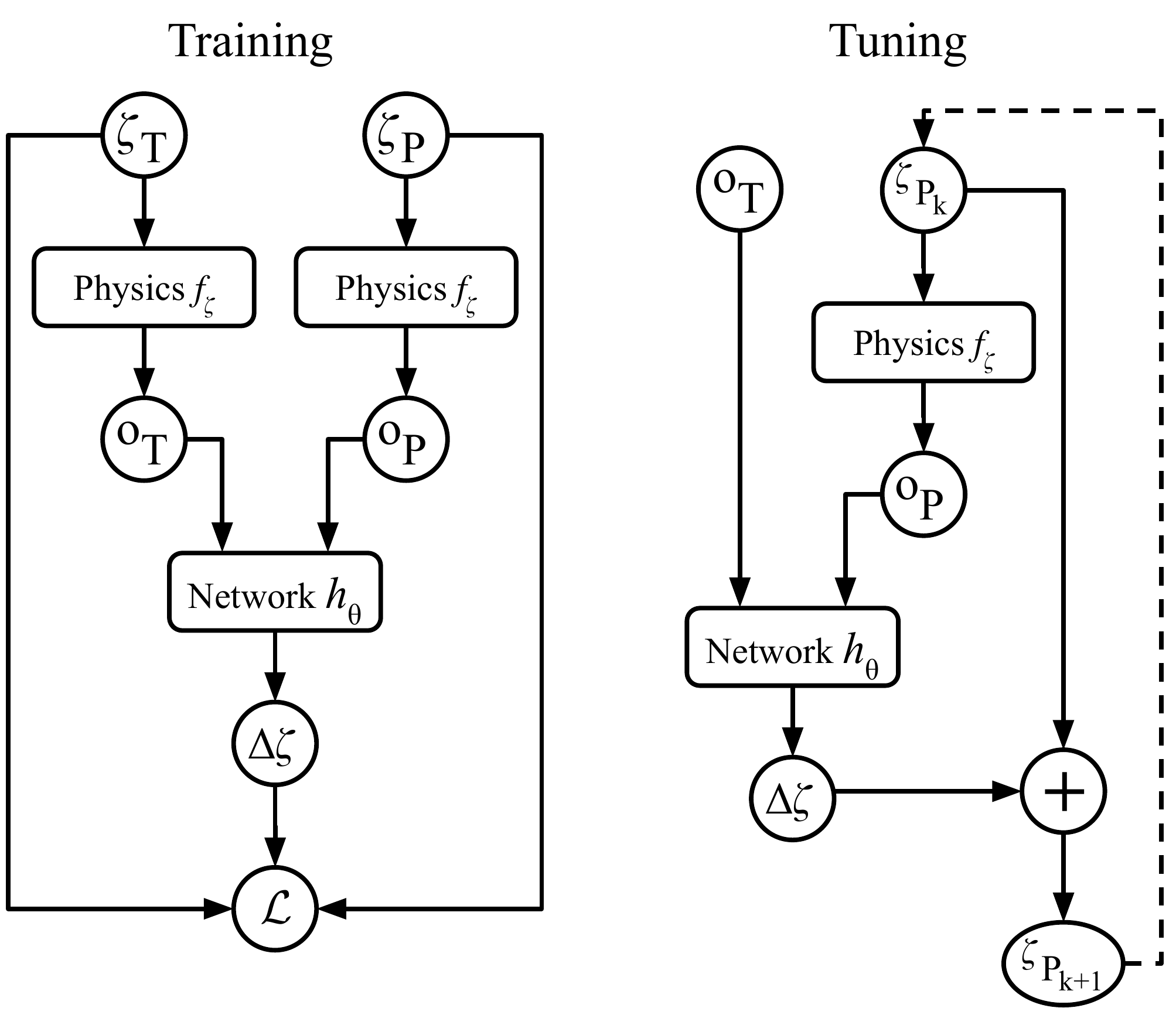}
    \hspace{1em}
    \includegraphics[width=0.4\textwidth]{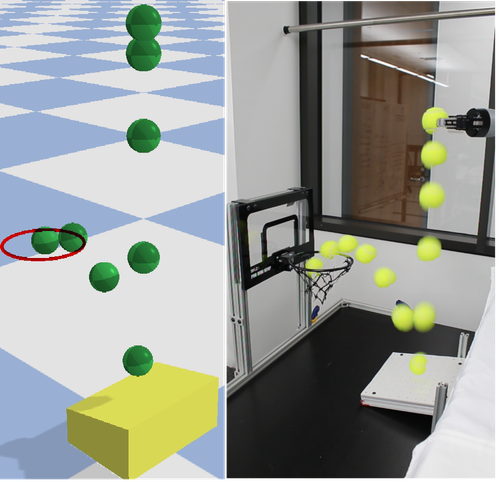}
    \caption{TuneNet uses observations to narrow the reality gap by tuning the physical parameters of a simulator so that it more closely approximates the real world (left), allowing sim-to-real robot skill transfer (right).}
    \label{fig:overview}
\end{figure}

Our primary contribution is the concept of \textit{residual tuning}, which allows fast and accurate one-shot system identification, and the development and analysis of the TuneNet neural network model for applying iterative parameter updates. We show that TuneNet is \textbf{fast} and \textbf{effective}: it tunes more efficiently than other gradient-free optimization methods, and that it outperforms state-of-the-art techniques for predicting rigid body motion. Finally, we validate that TuneNet can tune simulations to match real-world environments, and use those simulations to conduct sim-to-real task learning by teaching a robot to complete a task that requires accurate dynamics prediction.

\section{Related Work}

This paper focuses on tuning simulation parameters as a way to help rapidly close the reality gap. This places it at the intersection of system identification and sim-to-real learning.

Classical system identification \citep{khosla_1985_parameter_identification, gautier_1988_identifying_robot_parameters} used maximum likelihood methods to estimate the parameters of a fixed model given observed behavior. For estimating parameters of complex non-differentiable models (such as physics engines), robotics researchers have used a variety of gradient-free \citep{Chebotar2018SimToRealAdaptation, zhu_2018_vgmi, tan2016walking, zhu_2019_bayesian_sysid} or gradient-approximating \cite{Kolev2015PhysicallyConsistent} optimization techniques. %
These techniques require sampling the model (e.g. performing simulation rollouts) many times with different parameter values, which is a computationally expensive process. TuneNet seeks to minimize the number of simulation rollouts during system identification. Research has shown the ability to perform robot system identification from a few observations using a neural network \citep{yu2017up_osi}, but cannot take advantage of previous parameter estimations, and the results have not been validated in a sim-to-real setting.

In contrast to system identification, recent work (\citep{RajeswaranEPOpt, Chebotar2018SimToRealAdaptation}) has explored data-driven methods for simulation improvement by leveraging a combination of real-world data and domain randomization \citep{Tobin2017}. These adaptive approaches require huge numbers of randomized simulations to be repeated once the robot is in the target environment.

TuneNet also has similarities to techniques that learn a transformation or mapping from a simulated training environment to a target environment. These techniques leverage built-in priors in the form of physics simulators, while also providing a way to account for simulator inaccuracies. This may involve a neural network-based transformation over states \citep{golemo2018simrealtransfer} or actions \citep{AAAI17-Hanna}, learning a new real-world policy based on one learned in simulation \citep{rusu2017sim}, or using a combination of a physics model and a data-driven \textit{residual} to improve physics prediction \citep{ajay_2018_augmenting_simulators_bouncing, kloss2018_combined_models, zeng_2019_tossingbot} or policy learning \citep{Johannink_2018_residual, zeng_2019_tossingbot, silver_2018_residual}. All of these approaches require a substantial amount of real-world training examples from the target environment to learn a transformation or develop a new policy. In contrast, our approach learns to account for errors in parameter space rather than action or state space, and is pretrained in simulation, allowing a model to be adapted using a single observation from the target environment.

A recent related work is \citet{james_2018_sim2sim}, which trained a network to convert from randomized simulation images to non-randomized simulations, and showed that this network could also convert real-world pictures to simulations. \citet{zhang2019vr} took a similar approach, using adversarial domain adaptation \citep{hoffman_2017_cycada} and unlabeled data from the target environment to convert real images to simulated ones. Both of these works operate in the visual domain and focus on learning the arrangement of objects in an environment, whereas our approach focuses on learning better simulator physical parameters by observing the world state over time. Finally, \citet{xu_2019_densephysnet} discovered physics parameters through robot interaction, but the possible parameter values are fixed and the approach does not take advantage of information that could be gleaned from an existing simulation.

\section{One-Shot Residual Tuning}
In this section, we describe our approach for tuning a simulator to match real-world observations. We present the underlying theory first, followed by our algorithm and implementation.

\subsection{Tuning a Parameterized Model}
Our goal is to generate a \textit{proposed model}, $f_{\zeta_P}(s_t, a_t) = s_{t+1}$, so that it approximates another (fixed) \textit{target model}, $g_{\zeta_T}(s_t, a_t) = s_{t+1}$, where $\zeta_P$ and $\zeta_T$ are physics parameters of those models. One way to develop this approximation would be to minimize the prediction error (some distance function $d(\cdot, \cdot)$) over a dataset of $N$ examples (\cref{eqn:basic_loss}), where $L$ is the length of each example. 

\begin{equation}
    \label{eqn:basic_loss}
    \argmin_{\zeta_P} \sum_{n=1}^{N}{\sum_{t=1}^{L}{ d(f_{\zeta_P}(s_{n_t}, a_{n_t}),  g_{\zeta_T}(s_{n_t}, a_{n_t}))}}
\end{equation}

Unfortunately, we usually do not have access to the true system dynamics $g_{\zeta_T}$. In addition, if the proposed model is an off-the-shelf simulator, it is not differentiable, and so we cannot easily calculate a gradient for \cref{eqn:basic_loss}. While we can directly estimate the parameters using a neural network (as in \cite{yu2017up_osi}), the approach would not generalize well outside the training set, and would be unable to iteratively improve a parameter estimate without additional data since it would encode a single mapping from states to parameter values.

In this work, we instead assume that there exists some set of parameters $\zeta_T'$ that will cause the proposed model to closely match the target model, $f_{\zeta_T'} \approx g_{\zeta_T}$, and seek to approximate it. This assumption takes advantage of the fact that $f_\zeta$ encodes at least a reasonable approximation of of $g_\zeta$'s dynamics, and allows us to use the proposed model as a prior during optimization. %

\subsection{TuneNet}

Our model, TuneNet, estimates the \textit{parameter residual} $\Delta\zeta = \zeta_T' - \zeta_P \approx \zeta_T - \zeta_P$ between a single proposed model and a single target model, and then iteratively updates the proposed model parameters accordingly to match those of the target model. By using a network to estimate parameter deltas, we transform \cref{eqn:basic_loss} into a differentiable loss that can easily be used as a supervised learning objective. We can still calculate the quality of the world state estimation, but it does not appear in the loss.

We may be unable to directly observe the state of each model, and represent this with observation functions $z_P$ and $z_T$ applied to proposed and target models, respectively. Observations of length $L$ are generated from each model using the current proposed parameter $\zeta_P$ and the true target parameter $\zeta_T$ (\cref{eqn:observation_p} and \cref{eqn:observation_t}).

\begin{equation}
\label{eqn:observation_p}
o_P = z_P(f_{\zeta_P}(s_t, a_t)), t = 1\ldots L
\end{equation}
\begin{equation}
\label{eqn:observation_t}
o_T = z_T(g_{\zeta_T}(s_t, a_t)), t = 1\ldots L
\end{equation}

TuneNet estimates the parameter error directly based on the two observations and approximates a function $h$, which is parameterized by some $\theta$ (in this work, the weights of a neural network): $h_\theta(o_P, o_T) = \Delta\zeta$. The new optimization problem for training (over a dataset of length $n$) is given in \cref{eqn:final_loss}, where $\lambda$ is a weight regularization constant.
\begin{equation}
\label{eqn:final_loss}
\argmin_{\theta} \sum\limits_{n=1}^N \norm{\left( \zeta_{P_n} + h_\theta(o_{P_n}, o_{T_n}) \right) - \zeta_{T_n}} + \lambda \norm{\theta}^2
\end{equation}

Our network consists of one independent fully connected layer for each observation, the outputs of which are concatenated and passed to additional connected layers which estimate the parameter error. We found that concatenating the two intermediate outputs performed better that other combination methods, such as calculating the difference. In this work, we use a physics-based simulator for the proposed dynamics model $f_\zeta$, and the target model $g_\zeta$ is either another simulator or the real world.

We train TuneNet using a dataset where each datapoint consists of a \textit{pair} of models with a known difference in physics parameters (see \cref{fig:overview}). The datasets are auto-generated purely in simulation, which removes the need for hand-labeling or real-world data and makes training faster, safer, and easier. Also, because pairs are generated from the same dynamics model (differing only by $\zeta$), we know that during training, $\zeta_T = \zeta_T'$ and the best-case reality gap during training is 0.

After training TuneNet, we apply it to a set of observations from a new target model and observations from an initial proposed model having parameters $\zeta_{P}$ (\cref{fig:overview}). As discussed below in \cref{sec:theory}, TuneNet excels at estimating smaller differences in $\zeta$, so we tune iteratively in $K$ steps to achieve the final $\zeta_P = {\zeta_{P}}_{k=K}$ given a single target observation. The process can be described and implemented using a recursive approach, or an iterative one as listed in \cref{alg:algo}. The \textsc{Resample} function interpolates the values in $o_T$, if necessary, so that the length and sampling rate match those of $o_P$. The two observations are then passed to TuneNet to estimate the parameter difference between them. The algorithm adds the resulting estimate to the previous best guess for $\zeta_P$, bringing it into closer agreement with $\zeta_T$:

\begin{equation}
\label{eqn:update}
\zeta_{P_{k}} = \zeta_{P_{k-1}} + h_\theta(o_{P_{k-1}}, o_T), k=1\ldots K
\end{equation}

Importantly, we are able to do this without collecting any additional observations from the target model, and only sampling the proposed model once per iteration (\cref{eqn:update}). After tuning, the tuned simulator $f_{\zeta_P}$ is ready to be used for other tasks such as object motion prediction and robot task planning. 

\begin{algorithm}[t]
\caption{TuneNet's parameter tuning procedure.}
\label{alg:algo}
\begin{algorithmic}
\Procedure{Tune}{$o_T$, simulator $f_{\zeta}$, initial guess $\zeta_{P_0}$, Network $h_\theta$, $K$}
    \For{$k=1\ldots K$}
        \State $o_{P} \gets z_P(f_{\zeta_{P_{k-1}}}(s_t, a_t)), t = 1\ldots L$ \Comment{\cref{eqn:observation_p}}
        \State $o_T \gets \textproc{Resample}(o_T, L)$
        \State $\zeta_{P_k} \gets \zeta_{P_{k-1}} + h_\theta(o_P, o_T)$ \Comment{\cref{eqn:update}}
    \EndFor
    \State \textbf{return} $\zeta_{P_K}$
\EndProcedure
\end{algorithmic}
\end{algorithm}

\subsection{Residual Tuning}
\label{sec:theory}

By learning to estimate $\Delta\zeta$ instead of $\zeta$, an estimator naturally becomes better at estimating small deltas due to the nature of the difference transformation. Given uniformly distributed training data over some interval $[d, u]$, the probability densities of parameters $\zeta_P$ and $\zeta_T$ are equal to some constant $c = \frac{1}{u-d}$ over the interval. However, the difference, $\Delta\zeta$, will be distributed according to $P(\Delta\zeta = x) = -\frac{2}{(u-d)^2}x+\frac{2}{u-d}$, making small delta values more common. As the parameter estimate becomes closer and closer to the true value, the network is able to leverage more and more training data. This behavior allows us to conduct \textit{iterative residual tuning} by collecting observations from the proposed model using successively better ${\zeta_P}$ estimates. Estimating parameter deltas also allows TuneNet to tune to values that lie outside the ranges seen in training data, as opposed to other neural network approaches that often require a transformation or fine-tuning for a new output range.

Residual tuning can be seen as gradient descent in parameter space, using a neural network to approximate the gradient. While gradient-based optimization techniques such as momentum could potentially be combined with this approach, in this paper we show that TuneNet performs well using standard gradient descent.

\section{Experiments}
We performed three experiments to test that TuneNet can efficiently tune an off-the-shelf simulator to match a target environment, including ones it has not encountered before (Exp. 1 and 2), that it can generate tuned simulations that can be used for object motion prediction (Exp. 2), that it can be used to estimate real-world parameters, even when trained purely in simulation (Exp. 3), and that we can use a simulator tuned with TuneNet as a platform for learning tasks different than the one used for data collection (Exp. 3). We implement all networks using PyTorch \cite{pytorch2017}. Network architecture details, layer sizes, and training hyperparmaeters are provided in the appendix.

\subsection{Experiment 1: System Identification in Simulation}
We first performed a simulated experiment to validate TuneNet's ability to tune one model to match another. The tuned parameter was the mass of an object held in a robot's gripper---a classic system identification objective. To generate training data for the experiment, we use pairs of PyBullet\footnote{\href{https://pybullet.org}{https://pybullet.org}} simulations. Each simulation consists of a Kinova Jaco 7-DOF robot arm with a Robotiq 85 2-finger gripper at the end effector, moving in a circle as it holds an object (see appendix for task details). During this motion, we record the 7 raw joint torques over time, and provide this as the input observation signal. To make this task more difficult, in the test set, the \textit{target} simulator has its object mass increased by an additional \SI{1}{kg}, putting the all correct mass values completely outside the range of values seen during training. This experiment therefore evaluates TuneNet's ability to complete more general tuning tasks, and ones that may take multiple iterations to complete.

After training, we applied TuneNet with $K=9$ iterations over the validation and test sets to measure the estimation accuracy ($K$ was chosen empirically based on observed tuning behavior). We report the mean absolute error (MAE) of the estimated object mass over both the validation and test sets. We compare TuneNet's performance to several baselines: a prediction of \SI{1}{kg} for every data point (which would be the best our network could do if it was unable to tune values outside those seen in the training set) and a constant prediction of the mean parameter value in the \textit{test} set. %

On this task, TuneNet achieved an MAE of $\pm\SI{0.011}{kg}$ over the validation set, where both target and proposed masses were were in the training range (see \cref{tbl:mass_tuning_results}). When evaluating on the test set, where the target masses were all outside the range used for training, %
TuneNet still achieved an MAE of $\pm\SI{0.198}{kg}$---higher than on the validation set, but still far better than either baseline of always predicting \SI{1}{kg} (MAE=$\SI{0.504}{kg}$) or predicting the average test-set value (MAE=$\SI{0.255}{kg}$). %

\begin{table}[]
\begin{minipage}{\columnwidth}
\centering
\caption{Percent and absolute error values on the end effector mass system identification task.}

\label{tbl:mass_tuning_results}
\resizebox{0.8\textwidth}{!}{%
\begin{tabular}{llll}
\toprule
Method    & Val Set (Absolute) & Test Set (Absolute) & Test Set (Mean \%)\\
\midrule
Max of training range (1kg) & --             & 0.504   &  28.0\%\\
Mean of test range (1.504kg)  & 0.25           & 0.255    &  17.1\%\\
TuneNet   & \textbf{0.011}          & \textbf{0.198}    &  13.2\%\\
\bottomrule
\end{tabular}
}
\end{minipage}

\end{table}

\subsection{Experiment 2: Bouncing Ball in Simulation}
To further test TuneNet's ability to tune parameters, as well as its ability to predict rigid-body motion, we used it to tune the coefficient of restitution (COR) of a simulated bouncing ball. Our proposed and target dynamics models were again PyBullet simulations. For training, we recreated the simulated dataset used in the work of \citet{ajay_2018_augmenting_simulators_bouncing}: in each episode, a 0.5m-radius sphere is dropped onto a flat plane from a starting height in the range 4-5m, having $\text{COR}\sim\text{U}(0.3, 0.7)$. At each timestep, we recorded the $xyz$ position of the ball as the world state. We used the same sampling and dataset parameters as \citep{ajay_2018_augmenting_simulators_bouncing}---800 simulations for training and 100 for testing, with a physics update rate of 60Hz and episode length of 400 frames. We also note that unlike in \citep{ajay_2018_augmenting_simulators_bouncing}, each example in our dataset consists of information for a pair of simulations. In this experiment, the observation functions $z_P$ and $z_T$ were identity, i.e., we had direct access to the world state ground truth. We refer to this case as \textit{GT} (ground truth) in the rest of the paper.

We test over two datasets, GT Easy and GT Hard. GT Easy consists of simulations with proposed and target COR values in the range $[0.3, 0.7]$, the same as in the training set (this is for a more fair comparison with \citep{ajay_2018_augmenting_simulators_bouncing}). In GT Hard, the target COR values are extended to the range $[0, 1.0]$.

After training, we evaluated TuneNet by applying $K=5$ tuning iterations to tune a unique model for each run in the test set. For each run, we report the mean absolute error (MAE) between the final proposed model's COR and the target model's COR. We compare against four baselines: predicting the mean of the \textit{test} set target parameters (\textit{Mean}), our implementation of greedy entropy search \citep{zhu_2018_vgmi} (\textit{EntSearch}), the \textit{CMA-ES} optimization method \citep{hansen_2016_cmaes}, and a neural network that learns to directly predict parameters in one step (\textit{direct prediction}). Details on baseline implementations are in the appendix. To test our tuned model's overall accuracy in object motion prediction, we record the ball's position from a rollout of the final proposed model, compare it to the target model, and report the MAE and percent error (trans err, cm, and trans err \%) over all runs and timesteps.

\begin{figure}[t]
\centering
\includegraphics[width=0.48\textwidth]{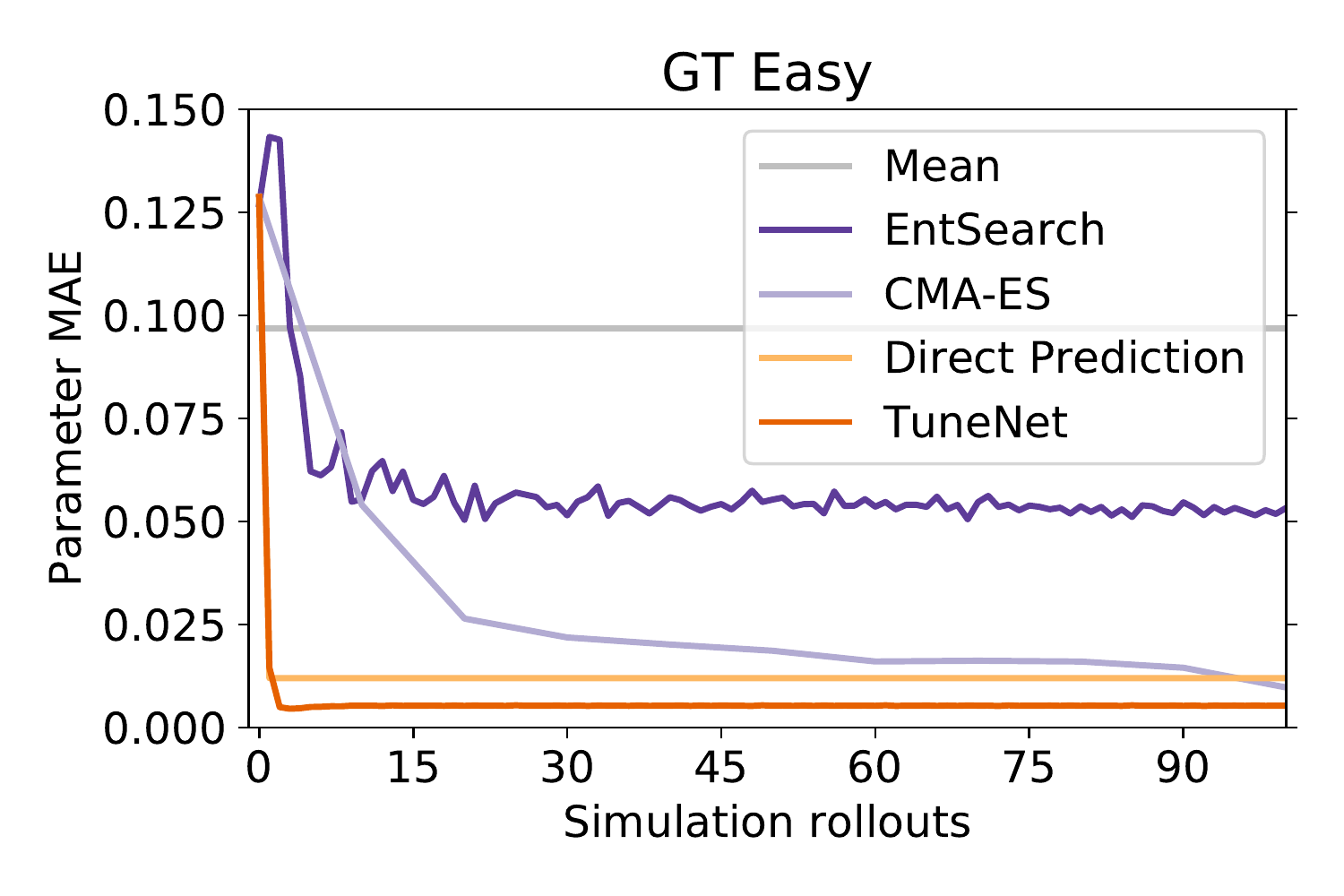} \includegraphics[width=0.48\textwidth]{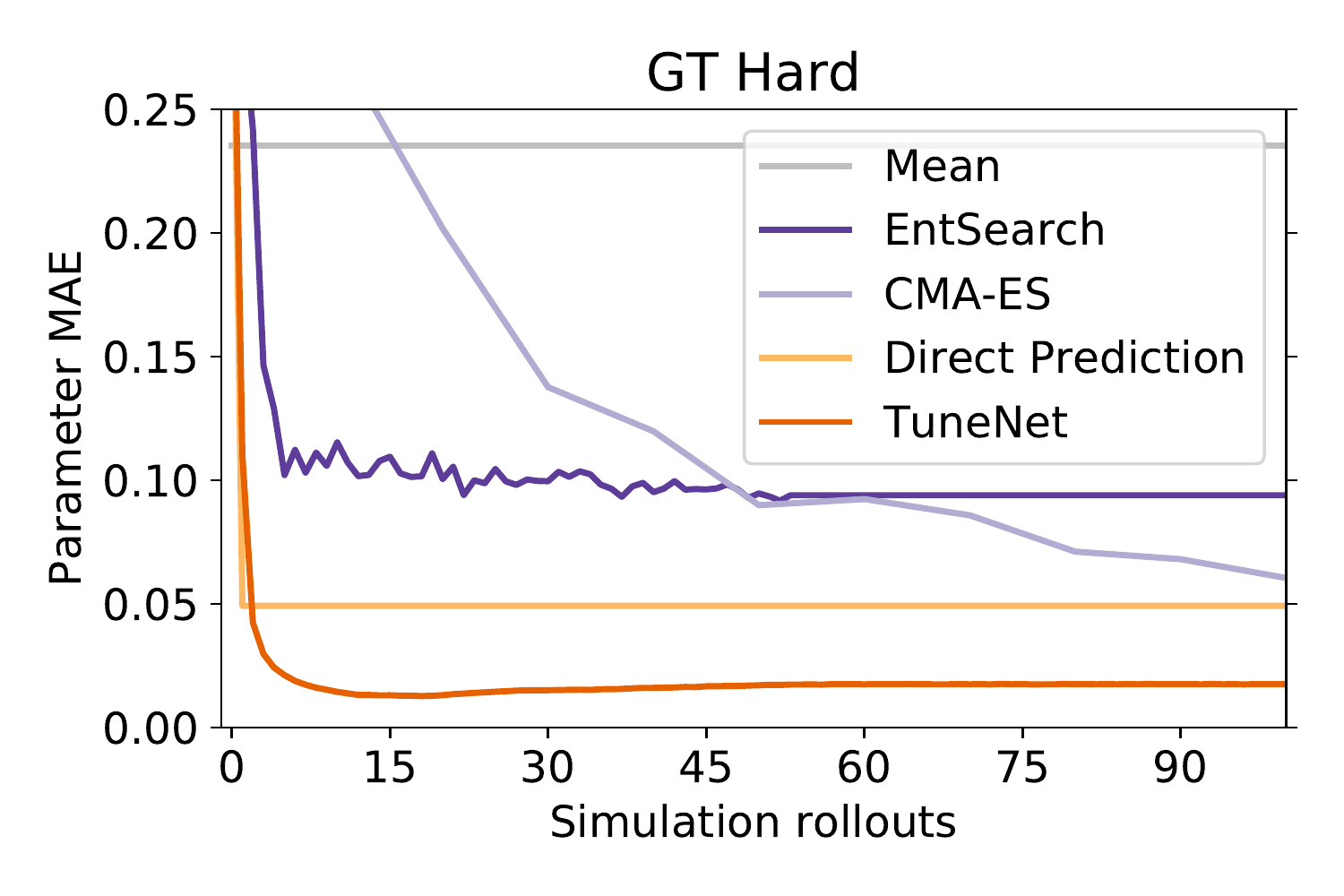}
\caption{Mean parameter error after $N$ iterations over the simulated ball-bouncing test set for various estimation techniques. TuneNet achieves the lowest error using just a few simulated iterations. Left: GT Easy. Right: GT Hard.}
\label{fig:iterations_vs_performance}
\end{figure}

\begin{table}[t]
\caption{Mean absolute error in COR prediction for the simulated ball-bouncing test set at various numbers of simulation rollouts. Best results in bold.}
\begin{minipage}{\columnwidth}
\centering
\begin{tabular}{l|llll|llll}
\toprule
  & \multicolumn{4}{c|}{GT Easy} & \multicolumn{4}{c}{GT Hard}                                    \\
Method       & $K=1$ & 5 & 10 & 100 & $K=1$ & 5 & 10 & 100 \\ \midrule
Mean                 & 0.0968 & 0.0968 & 0.0968 & 0.0968 & 0.2353 & 0.2353 & 0.2353 & 0.2353 \\
EntSearch            & 0.1433 & 0.0622 & 0.0553 & 0.0533 & 0.2876 & 0.1021 & 0.1153 & 0.0939 \\
CMA-ES               & 0.0540 & 0.0540 & 0.0540 & 0.0097 & 0.2766 & 0.2766 & 0.2766 & 0.0605 \\
Direct Prediction    & \textbf{0.0120} & 0.0120 & 0.0120 & 0.0120 & \textbf{0.0493} & 0.0493 & 0.0493 & 0.0493 \\
TuneNet              & 0.0145 & \textbf{0.0050} & \textbf{0.0053} & \textbf{0.0053} & 0.1103 & \textbf{0.0212} & \textbf{0.0145} & \textbf{0.0175} \\
\bottomrule
\end{tabular}
\end{minipage}
\label{tbl:mae_cor}
\end{table}

\cref{fig:iterations_vs_performance} compares the average tuning performance (measured by the MAE between the true and estimated parameters) as more and more simulation rollouts are used for estimation. \textit{Direct Prediction} and \textit{Mean} require 0 rollouts, since they are computed directly from the target observation. \textit{CMA-ES} uses 10 rollouts per iteration, and the other algorithms use 1 rollout per iteration. \cref{tbl:mae_cor} quantifies the values in the table for various numbers of simulation rollouts, averaged across the entire training set. TuneNet's one-step prediction accuracy is slightly lower than direct prediction, but it achieves the best accuracy of all methods in less than 5 iterations. In contrast, \textit{EntSearch} has difficulty converging to a good value, and \textit{CMA-ES} requires far more samples to develop a good estimate.

\begin{table}[t]
\caption{Mean absolute error in object motion prediction for the GT Easy dataset (Exp. 2).}
\begin{minipage}{\columnwidth}
\centering

\resizebox{0.45\textwidth}{!}{%
\begin{tabular}{@{}lccc@{}}
\toprule
Method          & trans err, \% & trans err, cm  \\ \midrule
Zero (inelastic)       & 100         & 97.0                        \\
Un-Tuned Physics        & 20.57          & 13.8                         \\
Neural \cite{ajay_2018_augmenting_simulators_bouncing}         & 9.16           & 5.8                          \\
VRNN Hybrid \cite{ajay_2018_augmenting_simulators_bouncing}    & 2.42           & 1.6                          \\
TuneNet (Obs)  & 6.03 & 5.7 \\
TuneNet (GT)  & \textbf{1.08} & \textbf{0.7} \\
\bottomrule
\end{tabular}
}
\end{minipage}
\label{tbl:mae_physics}
\end{table}

\cref{tbl:mae_physics} shows the object motion prediction error for the bouncing ball dataset. We compare to \citet{ajay_2018_augmenting_simulators_bouncing}'s VRNN hybrid network, although we note that those results are from one-step prediction models, whereas our model uses the entire state history to tune and is reported averaged over the entire dataset. TuneNet GT outperforms all other approaches for position prediction, with an MAE of \SI{7}{mm}, or 1.08\%. %

\subsection{Experiment 3: Tuning for Sim-to-Real Task Learning}
\begin{figure}[t]
    \begin{centering}
        \begin{subfigure}[t]{0.55\textwidth}
            \centering
            \begin{minipage}[t]{\textwidth}
                \centering
                \includegraphics[width=\textwidth]{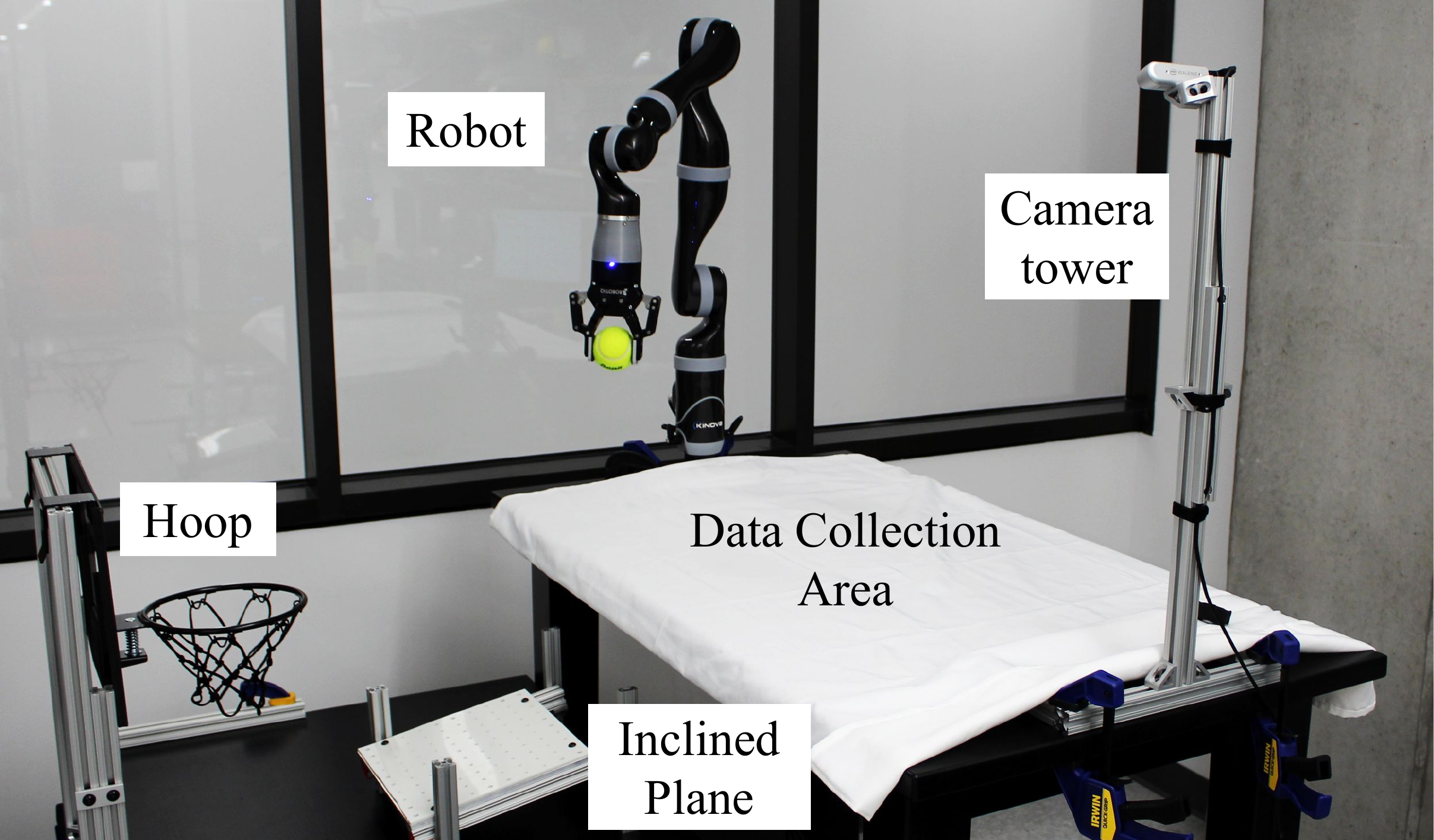}
            \end{minipage}
            \caption{}
            \label{fig:robot_setup}
        \end{subfigure}
        \hfill
        \begin{subfigure}[t]{0.44\textwidth}
            \centering
            \includegraphics[width=0.9\textwidth]{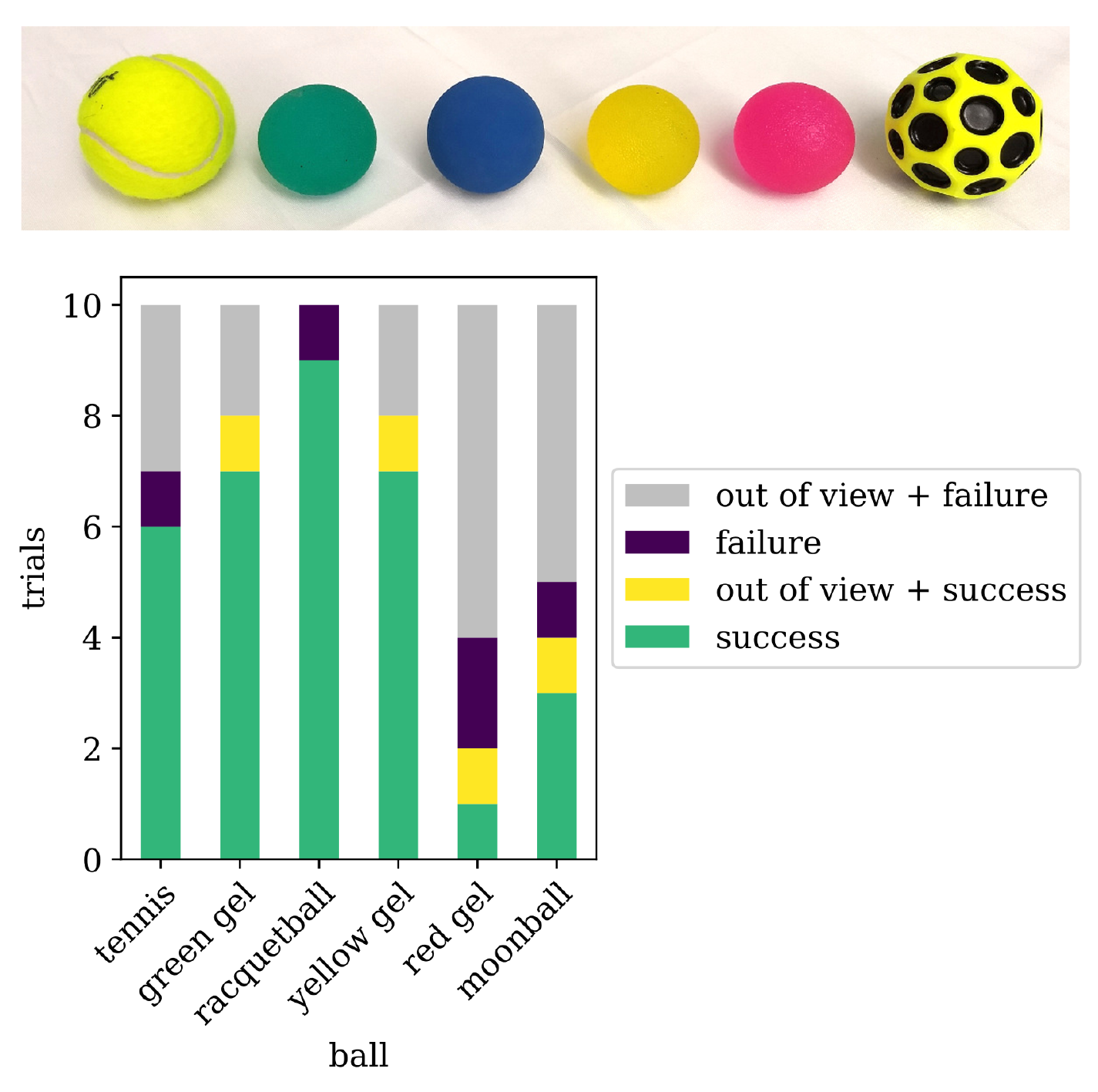}
            \caption{}
            \label{fig:bounce_success_rates}
        \end{subfigure}
        \begin{subfigure}[t]{\textwidth}
            \centering
            \begin{minipage}[t]{\textwidth}
                \centering
                \includegraphics[width=\textwidth]{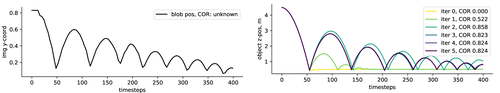}
            \end{minipage}
            \caption{}
            \label{fig:tuning_results}
        \end{subfigure}
    \end{centering}
    \caption{Experiment 3 sim-to-real task results. (a): The hardware setup used to collect data and perform tasks. (b): The six ball types used for evaluation and the bounce shot success rates for each type. (c) TuneNet's iterative tuning procedure. Left: observations from uncalibrated RGB video of robot dropping the \textit{red gel} ball (measured COR: 0.791). Right: bounce trajectories from a simulator using 5 TuneNet iterations to tune the ball's coefficient of restitution (COR), starting from an initial guess of COR=0 (no bouncing).}
    \label{fig:robot_task_results}
\end{figure}

Finally, we tested TuneNet for sim-to-real learning by estimating the COR of a bouncing ball again, but this time in the real world. Our task is to use a tuned simulator to bounce balls off an inclined plane into a hoop (a "bounce shot"), as seen in \cref{fig:robot_setup}. This is a difficult task, as striking the rim of the hoop will often cause the ball to bounce away rather than passing through the hoop. 

Our hardware platform for evaluation consists of a Kinova Jaco 7DOF robot arm with a Robotiq 85 2-finger gripper (see \cref{fig:robot_setup}), controlled using ROS\footnote{\href{https://ros.org}{https://ros.org}}. A camera is positioned opposite the robot for perception purposes.%

We trained TuneNet on a visual version of the GT Easy dataset used in Experiment 2. For the \textit{target} simulator in each episode, we rendered a 2D video using PyBullet's renderer, and applied an RGB-only OpenCV object tracker to calculate the position of the ball in each frame. The tuning problem remains challenging after using this object detector, since in addition to the presence of visual noise, 2D pixel coordinates do not directly correspond to z-height in 3 dimensions and our camera images are uncalibrated. The network input for each tuning task is the ground truth 3D position for the proposed model and the visually-derived 2D pixel position for the target model. This helps test TuneNet's ability to generalize and learn different transformations for each input channel. We refer to this network as \text{TuneNet Obs} in the rest of the paper.

For comparison purposes, we report TuneNet Obs performance in object prediction over the simulated dataset from Experiment 2 (\cref{tbl:mae_physics}). It still achieves position accuracy within 6\%, and while it performs more poorly than the GT case, this is expected, since the network is operating on uncalibrated and noisy visual observations and the other approaches in the table have access to the true simulator state.

For the bounce shot task, we performed 10 independent tuning trials for 6 different ball types. In each trial, the robot first dropped the ball on a flat table and recorded a video of the resulting bounces. The video was resampled to 60 Hz, clipped to 400 frames, and processed using the same OpenCV object tracker as on the training set. TuneNet operated on the observation from one real-world drop and the ground truth state from an un-tuned simulator which started with a COR of 0 (no bounce). We used TuneNet ($K=5$) to tune the simulated COR, clamping to the range $[0, 1]$, as values outside this range are physically meaningless. 

\cref{fig:tuning_results} shows an example of TuneNet using $K=5$ iterative tuning rounds to match observations from a video of a dropped racquetball, beginning with a COR of 0. The first parameter updates are large, and successive iterations fine-tune the estimate because of the more accurate tuning behavior for small parameter deltas (see \cref{sec:theory}).

The tuned COR value was used to construct a new simulator, consisting of a dropped ball, inclined plane, and hoop that match our real-world task setup. The simulator was used to uniformly sample 100 shot heights from a range within the robot's reachable space, and the optimal height was chosen as the shot height that passed the ball as close to the center of the hoop as possible. The real robot then executes the shot from the optimal height. We repeat the task for all 10 independent trials and all 6 ball types (not all of which are round, see \cref{fig:robot_task_results}) and record the success or failure of the bounce shot.

\cref{fig:robot_task_results} shows the results of the bounce shot task and a representative execution. Overall, the robot succeeded at 62\% of the bounce shots. The primary reason for failure was that during the pre-tuning data collection step, the ball bounced or rolled off the table or out of view of the camera. This resulted in a meaningless input observation, which caused in a low estimated COR and incorrect drop height. Indeed, when these "off table" trials are removed, the success rate increases to 87\%, whereas the "off table" trials were only 18\% successful, as shown in \cref{fig:bounce_success_rates}. Eliminating these examples where the ball bounced off the table in the observation drop, two of the six ball types were successful in every trial. Successful shots with all 6 balls can be found in the supplementary video.

\section{Discussion and Conclusion}
\label{sec:discussion}

TuneNet is \textbf{fast}. Since each tuning iteration consists of a single network forward pass and a single simulation rollout, TuneNet reduces the number of parameter-space simulation samples required compared to existing gradient-free optimization, and it tunes significantly faster than techniques that alternate between domain randomization in simulation and real-world data collection \citep{zhu_2018_vgmi, Chebotar2018SimToRealAdaptation}. Our dataset generation and training procedure is also fast due to being trained entirely in simulation, and completes in under 7 minutes using a single NVIDIA GTX 1080Ti. 

TuneNet is \textbf{effective}. Our results show that even though a simulation cannot perfectly model the real world, much of the reality gap can be closed by choosing good simulation parameters. We also note that TuneNet is compatible with recent approaches that learn data-driven transformations or residuals to improve simulations \cite{kloss2018_combined_models, hanna_stone_2017_behavior_policy_search, golemo2018simrealtransfer, zeng_2019_tossingbot}, and envision a two-step approach where TuneNet first matches a real-world environment as closely as possible using parameter tuning alone, then additional unmodeled effects are accounted for using learned transformations or domain randomization.

That said, there are some limitations of our approach that suggest avenues for future work. For example, parameters may not be fully identifiable based on observations alone, as explored in several studies \citep{ayusawa_2013_underactuated_sysid, ma2018friction, fazeli2018identifiability}. As discussed in \citep{fazeli2018identifiability} and \citep{khosla_1985_parameter_identification}, the \textit{identifiable set} of parameters may be coupled, which would result in multiple local minima in parameter space. This may prove to be an issue if the goal is to accurately measure system parameters, but as long as the tuned parameters enable accurate prediction and sim-to-real learning using the proposed model, which local minimum the algorithm converges to is less important. For example, in our study, the \textit{racquetball} had high task success rates despite a technically inaccurate COR estimate ($0.789\pm0.024$ vs. regulation values of $0.68$--$0.72$ \cite{racquetball_rules}). We also note that \cite{ramos_2019_bayes_sim} is concurrently studying the problem of modeling belief over coupled parameter values.

In this paper, we introduced TuneNet, a neural network-based technique to perform residual tuning between two models, and validated the approach for use in system identification and sim-to-real robotics tasks. TuneNet represents a new method for closing the reality gap, showing that a simulator can be matched to the real world with just one observation and minimal simulation sampling. We plan to use TuneNet to rapidly generate tuned simulators for new environments, developing strong models for object manipulation without extensive real-world practice.

\section*{Acknowledgments}
We thank Josiah Hanna and Scott Niekum for their helpful suggestions on this paper. This work was supported by the National Science Foundation (IIS-1564080, IIS-1724157) and Office of Naval Research (N000141612835, N000141612785).

\bibliography{references_clean}

\begin{thebibliography}{32}
\providecommand{\natexlab}[1]{#1}
\providecommand{\url}[1]{\texttt{#1}}
\expandafter\ifx\csname urlstyle\endcsname\relax
  \providecommand{\doi}[1]{doi: #1}\else
  \providecommand{\doi}{doi: \begingroup \urlstyle{rm}\Url}\fi

\bibitem[Johannink et~al.(2018)Johannink, Bahl, Nair, Luo, Kumar, Loskyll,
  Ojea, Solowjow, and Levine]{Johannink_2018_residual}
T.~Johannink, S.~Bahl, A.~Nair, J.~Luo, A.~Kumar, M.~Loskyll, J.~A. Ojea,
  E.~Solowjow, and S.~Levine.
\newblock {Residual Reinforcement Learning for Robot Control}.
\newblock \emph{CoRR}, abs/1812.0, 2018.

\bibitem[Chebotar et~al.(2018)Chebotar, Handa, Makoviychuk, Macklin, Issac,
  Ratliff, and Fox]{Chebotar2018SimToRealAdaptation}
Y.~Chebotar, A.~Handa, V.~Makoviychuk, M.~Macklin, J.~Issac, N.~Ratliff, and
  D.~Fox.
\newblock {Closing the Sim-to-Real Loop: Adapting Simulation Randomization with
  Real World Experience}.
\newblock \emph{CoRR}, 10 2018.

\bibitem[Hanna and Stone(2017)]{AAAI17-Hanna}
J.~P. Hanna and P.~Stone.
\newblock {Grounded Action Transformation for Robot Learning in Simulation}.
\newblock In \emph{Proceedings of the Thirty-First AAAI Conference on
  Artificial Intelligence (AAAI-17)}, pages 3834--3840, 2 2017.
\newblock \doi{10.1016/J.OPTLASTEC.2016.08.019}.

\bibitem[Golemo et~al.(2018)Golemo, Ta{\"{i}}ga, Oudeyer, and
  Courville]{golemo2018simrealtransfer}
F.~Golemo, A.~A. Ta{\"{i}}ga, P.-Y. Oudeyer, and A.~Courville.
\newblock {Sim-to-Real Transfer with Neural-Augmented Robot Simulation}.
\newblock In \emph{2nd Conference on Robot Learning (CoRL18)}, volume~87, pages
  817--828, 2018.
\newblock URL \url{http://proceedings.mlr.press/v87/golemo18a/golemo18a.pdf}.

\bibitem[Zeng et~al.(2019)Zeng, Song, Lee, Rodriguez, and
  Funkhouser]{zeng_2019_tossingbot}
A.~Zeng, S.~Song, J.~Lee, A.~Rodriguez, and T.~A. Funkhouser.
\newblock {TossingBot: Learning to Throw Arbitrary Objects with Residual
  Physics}.
\newblock In \emph{Proceedings of Robotics: Science and Systems}, Freiburg im
  Breisgau, Germany, 6 2019.
\newblock \doi{10.15607/RSS.2019.XV.004}.

\bibitem[Silver et~al.(2018)Silver, Allen, Tenenbaum, and
  Kaelbling]{silver_2018_residual}
T.~Silver, K.~Allen, J.~Tenenbaum, and L.~P. Kaelbling.
\newblock {Residual Policy Learning}.
\newblock \emph{CoRR}, abs/1812.0, 2018.

\bibitem[Rusu et~al.(2017)Rusu, Ve{\v{c}}er{\'{i}}k, Roth{\"{o}}rl, Heess,
  Pascanu, and Hadsell]{rusu2017sim}
A.~A. Rusu, M.~Ve{\v{c}}er{\'{i}}k, T.~Roth{\"{o}}rl, N.~Heess, R.~Pascanu, and
  R.~Hadsell.
\newblock {Sim-to-Real Robot Learning from Pixels with Progressive Nets}.
\newblock In \emph{Conference on Robot Learning}, pages 262--270, 2017.

\bibitem[Tobin et~al.(2017)Tobin, Fong, Ray, Schneider, Zaremba, and
  Abbeel]{Tobin2017}
J.~Tobin, R.~Fong, A.~Ray, J.~Schneider, W.~Zaremba, and P.~Abbeel.
\newblock {Domain randomization for transferring deep neural networks from
  simulation to the real world}.
\newblock In \emph{IEEE International Conference on Intelligent Robots and
  Systems}, pages 23--30. IEEE, 9 2017.
\newblock \doi{10.1109/IROS.2017.8202133}.

\bibitem[Tan et~al.(2018)Tan, Zhang, Coumans, Iscen, Bai, Hafner, Bohez, and
  Vanhoucke]{tan_sim_to_real_2018}
J.~Tan, T.~Zhang, E.~Coumans, A.~Iscen, Y.~Bai, D.~Hafner, S.~Bohez, and
  V.~Vanhoucke.
\newblock {Sim-to-Real: Learning Agile Locomotion For Quadruped Robots}.
\newblock In \emph{Robotics: Science and Systems XIV}. Robotics: Science and
  Systems Foundation, 6 2018.
\newblock ISBN 1804.10332v2.
\newblock \doi{arXiv:1804.10332v2}.

\bibitem[Rajeswaran et~al.(2016)Rajeswaran, Ghotra, Ravindran, and
  Levine]{RajeswaranEPOpt}
A.~Rajeswaran, S.~Ghotra, B.~Ravindran, and S.~Levine.
\newblock {EPOpt: Learning Robust Neural Network Policies Using Model
  Ensembles}.
\newblock \emph{CoRR}, 2016.
\newblock ISSN 0027-8424.
\newblock \doi{10.1073/pnas.211563298}.

\bibitem[Pinto et~al.(2017)Pinto, Andrychowicz, Welinder, Zaremba, and
  Abbeel]{pinto_2018_asymmetric_rl}
L.~Pinto, M.~Andrychowicz, P.~Welinder, W.~Zaremba, and P.~Abbeel.
\newblock {Asymmetric Actor Critic for Image-Based Robot Learning}.
\newblock In \emph{Robotics: Science and Systems XIV}. Robotics: Science and
  Systems Foundation, 6 2017.
\newblock \doi{10.1007/s10869-008-9083-z}.

\bibitem[Khosla and Kanade(1985)]{khosla_1985_parameter_identification}
P.~Khosla and T.~Kanade.
\newblock {Parameter identification of robot dynamics}.
\newblock In \emph{1985 24th IEEE Conference on Decision and Control}, pages
  1754--1760. IEEE, 12 1985.
\newblock ISBN 9684641354.
\newblock \doi{10.1109/CDC.1985.268838}.

\bibitem[Gautier and Khalil(1988)]{gautier_1988_identifying_robot_parameters}
M.~Gautier and W.~Khalil.
\newblock {On the identification of the inertial parameters of robots}.
\newblock In \emph{Proceedings of the 27th IEEE Conference on Decision and
  Control}, pages 2264--2269. IEEE, 1988.
\newblock ISBN 0-8186-0852-8.
\newblock \doi{10.1109/CDC.1988.194738}.

\bibitem[Zhu et~al.(2018)Zhu, Kimmel, Bekris, and Boularias]{zhu_2018_vgmi}
S.~Zhu, A.~Kimmel, K.~E. Bekris, and A.~Boularias.
\newblock {Fast Model Identification via Physics Engines for Data-efficient
  Policy Search}.
\newblock In \emph{Proceedings of the 27th International Joint Conference on
  Artificial Intelligence (IJCAI)}, pages 3249--3256, 2018.
\newblock URL \url{http://dl.acm.org/citation.cfm?id=3304889.3305112}.

\bibitem[Tan et~al.(2016)Tan, Xie, Boots, and Liu]{tan2016walking}
J.~Tan, Z.~Xie, B.~Boots, and C.~K. Liu.
\newblock {Simulation-based design of dynamic controllers for humanoid
  balancing}.
\newblock In \emph{IEEE International Conference on Intelligent Robots and
  Systems}, pages 2729--2736. IEEE, 10 2016.
\newblock \doi{10.1109/IROS.2016.7759424}.

\bibitem[Zhu et~al.(2019)Zhu, Surovik, Bekris, and
  Boularias]{zhu_2019_bayesian_sysid}
S.~Zhu, D.~Surovik, K.~E. Bekris, and A.~Boularias.
\newblock {Closing the Reality Gap of Robotic Simulators through Task-oriented
  Bayesian Optimization}.
\newblock \emph{Journal of Machine Learning Research}, 2019.

\bibitem[Kolev and Todorov(2015)]{Kolev2015PhysicallyConsistent}
S.~Kolev and E.~Todorov.
\newblock {Physically consistent state estimation and system identification for
  contacts}.
\newblock In \emph{IEEE-RAS International Conference on Humanoid Robots},
  volume 2015-Decem, pages 1036--1043. IEEE, 11 2015.
\newblock ISBN 9781479968855.
\newblock \doi{10.1109/HUMANOIDS.2015.7363481}.

\bibitem[Yu et~al.(2017)Yu, Tan, Liu, and Turk]{yu2017up_osi}
W.~Yu, J.~Tan, C.~K. Liu, and G.~Turk.
\newblock {Preparing for the Unknown: Learning a Universal Policy with Online
  System Identification}.
\newblock In \emph{Proceedings of Robotics: Science and Systems}, Cambridge,
  Massachusetts, 7 2017.
\newblock \doi{10.15607/RSS.2017.XIII.048}.

\bibitem[Ajay et~al.(2018)Ajay, Wu, Fazeli, Bauza, Kaelbling, Tenenbaum, and
  Rodriguez]{ajay_2018_augmenting_simulators_bouncing}
A.~Ajay, J.~Wu, N.~Fazeli, M.~Bauza, L.~P. Kaelbling, J.~B. Tenenbaum, and
  A.~Rodriguez.
\newblock {Augmenting Physical Simulators with Stochastic Neural Networks: Case
  Study of Planar Pushing and Bouncing}.
\newblock In \emph{International Conference on Intelligent Robots and Systems
  (IROS)}, 2018.

\bibitem[Kloss et~al.(2017)Kloss, Schaal, and Bohg]{kloss2018_combined_models}
A.~Kloss, S.~Schaal, and J.~Bohg.
\newblock {Combining learned and analytical models for predicting action
  effects}.
\newblock \emph{CoRR}, 2017.

\bibitem[James et~al.(2019)James, Wohlhart, Kalakrishnan, Kalashnikov, Irpan,
  Ibarz, Levine, Hadsell, and Bousmalis]{james_2018_sim2sim}
S.~James, P.~Wohlhart, M.~Kalakrishnan, D.~Kalashnikov, A.~Irpan, J.~Ibarz,
  S.~Levine, R.~Hadsell, and K.~Bousmalis.
\newblock {Sim-to-Real via Sim-to-Sim: Data-efficient Robotic Grasping via
  Randomized-to-Canonical Adaptation Networks}.
\newblock In \emph{Computer Vision and Pattern Recognition (CVPR)}, 2019.

\bibitem[Zhang et~al.(2019)Zhang, Tai, Yun, Xiong, Liu, Boedecker, and
  Burgard]{zhang2019vr}
J.~Zhang, L.~Tai, P.~Yun, Y.~Xiong, M.~Liu, J.~Boedecker, and W.~Burgard.
\newblock {Vr-goggles for robots: Real-to-sim domain adaptation for visual
  control}.
\newblock \emph{IEEE Robotics and Automation Letters}, 4\penalty0 (2):\penalty0
  1148--1155, 2019.

\bibitem[Hoffman et~al.(2017)Hoffman, Tzeng, Park, Zhu, Isola, Saenko, Efros,
  and Darrell]{hoffman_2017_cycada}
J.~Hoffman, E.~Tzeng, T.~Park, J.-Y. Zhu, P.~Isola, K.~Saenko, A.~A. Efros, and
  T.~Darrell.
\newblock {CyCADA: Cycle-Consistent Adversarial Domain Adaptation}.
\newblock \emph{CoRR}, abs/1711.0, 2017.

\bibitem[Xu et~al.(2019)Xu, Wu, Zeng, Tenenbaum, and
  Song]{xu_2019_densephysnet}
Z.~Xu, J.~Wu, A.~Zeng, J.~Tenenbaum, and S.~Song.
\newblock {DensePhysNet: Learning Dense Physical Object Representations Via
  Multi-Step Dynamic Interactions}.
\newblock In \emph{Proceedings of Robotics: Science and Systems}, Freiburg im
  Breisgau, Germany, 6 2019.
\newblock \doi{10.15607/RSS.2019.XV.046}.

\bibitem[Paszke et~al.(2017)Paszke, Chanan, Lin, Gross, Yang, Antiga, and
  Devito]{pytorch2017}
A.~Paszke, G.~Chanan, Z.~Lin, S.~Gross, E.~Yang, L.~Antiga, and Z.~Devito.
\newblock {Automatic differentiation in PyTorch}.
\newblock In \emph{31st Conference on Neural Information Processing Systems
  (NIPS)}, pages 1--4, Long Beach, CA, 2017.
\newblock ISBN 9788578110796.
\newblock \doi{10.1017/CBO9781107707221.009}.

\bibitem[Hansen(2016)]{hansen_2016_cmaes}
N.~Hansen.
\newblock {The CMA Evolution Strategy: A Tutorial}.
\newblock \emph{CoRR}, abs/1604.0, 2016.

\bibitem[Hanna et~al.(2017)Hanna, Thomas, Stone, and
  Niekum]{hanna_stone_2017_behavior_policy_search}
J.~P. Hanna, P.~S. Thomas, P.~Stone, and S.~Niekum.
\newblock {Data-Efficient Policy Evaluation Through Behavior Policy Search}.
\newblock In \emph{Proceedings of the 34th International Conference on Machine
  Learning}, volume~70 of \emph{Proceedings of Machine Learning Research},
  pages 1394--1403, Sydney, Australia, 2017. PMLR.

\bibitem[Ayusawa et~al.(2013)Ayusawa, Venture, and
  Nakamura]{ayusawa_2013_underactuated_sysid}
K.~Ayusawa, G.~Venture, and Y.~Nakamura.
\newblock {Identifiability and identification of inertial parameters using the
  underactuated base-link dynamics for legged multibody systems}.
\newblock \emph{The International Journal of Robotics Research}, 33:\penalty0
  446--468, 2013.
\newblock \doi{10.1177/0278364913495932}.

\bibitem[Ma and Rodriguez(2018)]{ma2018friction}
D.~Ma and A.~Rodriguez.
\newblock {Friction Variability in Auto-collected Dataset of Planar Pushing
  Experiments and Anisotropic Friction}.
\newblock \emph{IEEE Robotics and Automation Letters}, 3\penalty0 (4):\penalty0
  3232--3239, 2018.
\newblock ISSN 2377-3766.
\newblock \doi{10.1109/LRA.2018.2851026}.

\bibitem[Fazeli et~al.(2015)Fazeli, Tedrake, and
  Rodriguez]{fazeli2018identifiability}
N.~Fazeli, R.~Tedrake, and A.~Rodriguez.
\newblock {Identifiability Analysis of Planar Rigid-Body Frictional Contact}.
\newblock In \emph{International Symposium on Robotics Research (ISRR)}, page
  To appear. Springer, 2015.
\newblock ISBN 9781424479740.

\bibitem[Dietrich(2015)]{racquetball_rules}
O.~E. Dietrich.
\newblock \emph{{USA Racquetball Official Rules of Racquetball}}.
\newblock USA Racquetball, 2015.
\newblock URL \url{https://www.teamusa.org/USA-Racquetball/How-To-Play/Rules/}.

\bibitem[Ramos et~al.(2019)Ramos, Possas, and Fox]{ramos_2019_bayes_sim}
F.~Ramos, R.~C. Possas, and D.~Fox.
\newblock {BayesSim: Adaptive Domain Randomization Via Probabilistic Inference
  for Robotics Simulators}.
\newblock In \emph{Proceedings of Robotics: Science and Systems}, Freiburg im
  Breisgau, Germany, 6 2019.
\newblock \doi{10.15607/RSS.2019.XV.029}.

\end{thebibliography}

\section*{Appendix}

\subsection*{TuneNet Architecture Implementation Details}
We implement all networks using PyTorch \cite{pytorch2017}, and train them using stochastic gradient descent. Our TuneNet architecture for all experiments consists of fully-connected (FC) layers with ReLU activation functions for each of the feature extractors, with each feature extractor output size set to 32. The estimator therefore has an input size of 64. We model the estimator with two fully connected layers with a hidden size of 32, ReLU activation for the first layer, and a \textit{tanh} activation for the final output. 

\subsection*{Details for Experiment 1: End-effector Mass Identification}
The object's mass is randomly chosen, $m \sim U(\SI{0}{kg}, \SI{1}{kg})$, for each training point. In each simulated episode, which runs for 5 seconds, the robot moves its end effector in a circle defined by the coordinates $[-0.4, 0.2\text{cos}(\tau), 0.2\text{sin}(\tau)], \tau = 1.2t$. We collect 1000, 300, and 300 episodes for the training, validation, and test sets, respectively. We trained the model for 1000 epochs using a batch size of 50, a learning rate of 0.01, L2 regularization $\lambda = \num{1e-3}$, and 1\% learning rate decay every 50 epochs. During training, the input torques were normalized to the range $[0, 1]$ but the outputs were not transformed. 

\subsection*{Details for Experiment 2: Bouncing Ball COR Tuning}
For both datasets, we trained TuneNet for 200 epochs, using a batch size of 50 and a learning rate of \num{1e-2}, with learning rate decay of 1\% per 5 epochs. The L2 normalization constant $\lambda$ is set to 0.01. During training, the input and output channels were normalized to the range $[0, 1]$. 

For the Obs case, the visual OpenCV tracker uses hue, saturation, and value windows to threshold a camera image and isolate an object. It then calculates the $y$ pixel position of the ball in each frame and normalizes the image pixel positions to the range $[0, 1]$. For each randomized simulation render, the PyBullet camera was placed at random polar coordinates $r \sim \text{U}(\SI{5}{m}, \SI{10}{m})$, $\theta \sim \text{U}(0, 2\pi)$, $z \sim \text{U}(\SI{5}{m}, \SI{8}{m})$, and aimed at the point $(0, 0, \SI{2}{m})$. 

\subsubsection*{Baseline Implementation Details}
For CMA-ES, we use the open-source package \texttt{pycma}\footnote{\url{https://github.com/CMA-ES/pycma}}. The inputs to CMA-ES are identical to those provided to TuneNet, with the initial proposed physics parameters being used as the initial guess for the CMA algorithm. The CMA tolerance was set to 0.1.

For Greedy Entropy Search (\textit{EntSearch} in this paper), we re-implemented the algorithm provided in \citep{zhu_2018_vgmi}. The underlying Gaussian Process (GP) sampling was implemented using the open-source Python package \texttt{GPy}\footnote{\url{http://github.com/SheffieldML/GPy}}, and we used an RBF GP kernel. Again, the initial proposed physics parameters were used as the initial guess for the algorithm. We discretize the parameter space into 50 bins, and use a gaussian process sampling population size of $n=100$. Because we do not calculate the value function for each policy, we cannot use the value as the stopping criterion as in \citet{zhu_2018_vgmi}. Instead, we stop iterating when the predicted parameter value does not change for two consecutive timesteps. This stopping parameter was determined empirically for our experiment after being shown to outperform two other stopping criteria: no stopping (which performed poorly because of noise in the GP sampling), and stopping after the parameter estimate changed by less than a threshold $\epsilon > 0$.

The initial guesses and simulated outcomes supplied to both CMA-ES and Greedy Entropy Search were identical to the inputs supplied to TuneNet, including all normalization transformations applied to the data.

\end{document}